\documentclass[runningheads,a4paper]{llncs}

\usepackage{amssymb}
\usepackage{graphicx}
\usepackage{algorithm}
\usepackage[noend]{algpseudocode}
\usepackage{graphicx}
\usepackage[table,xcdraw]{xcolor}
\usepackage{multirow}
\usepackage[export]{adjustbox}
\usepackage{url}
\usepackage[final]{pdfpages}

\urldef{\mailsa}\path|v.marochko@innopolis.ru|
\urldef{\mailsb}\path|l.johard@innopolis.ru|
\urldef{\mailsc}\path|m.mazzara@innopolis.ru|    
\newcommand{\keywords}[1]{\par\addvspace\baselineskip
\noindent\keywordname\enspace\ignorespaces#1}

\begin{document}

\mainmatter  
\title{Pseudorehearsal in value function approximation}

\titlerunning{Pseudorehearsal}

\author{Vladimir Marochko \and Leonard Johard \and Manuel Mazzara}

\institute{Innopolis University\\
Universitetskaya str. 1, 420500 Innopolis, Russia\\
}

\maketitle

\begin{abstract}
Catastrophic forgetting is of special importance in reinforcement learning, as the data distribution is generally non-stationary over time. We study and compare several pseudorehearsal approaches for Q-learning with function approximation in a pole balancing task. We have found that pseudorehearsal seems to assist learning even in such very simple problems, given proper initialization of the rehearsal parameters.

\end{abstract}

\section{Introduction}
Reinforcement learning is a more general problem formulation than the commonly used supervised learning framework. As such, it can be applied to a wider range of problems. It is also a more difficult problem to optimize for, as the feedback is more limited. 

Reinforcement learning covers space of prediction and control problems in partially unknown space with unknown behavior. The agent should explore the environment and find optimal actions it has to perform to reach the goal. It is used in cases when the optimal policy is unknown so there is no way to train agents using supervised learning algorithms. The basic idea can be described by the metaphor of a player starting an unknown game and, after a number of turns, he/she receives a message stating that he/she has lost or won. After a number of games he/she will figure out how to act to win as often as possible.

\subsection{Supervised agents in reinforcement learning problems}

In order to solve the practical problems that can be assumed to be approximately Markov decision process (MDP) \cite{murphy2000survey}, like robot's navigation, playing chess or trading on stock exchange, we can use a value function approximation to speed up the learning process \cite{johard2014connectionist} \cite{sutton1988learning}. The weakness of this approach is that convergence is not guaranteed if the MDP approximation is incorrect, or in cases where the inputs are continuous, and which necessitates non-linear function approximation \cite{tsitsiklis1997analysis}. Furthermore if we have continuous outputs the value approximation needs to be combined with an additional optimization technique, such as REINFORCE \cite{williams1992simple}, in order to search for optimal outputs. If the outputs are discrete, a simple maximization is sufficient and allows us to make use of the Q-learning framework \cite{watkins1992q}\cite{geist2010brief}.

Although a value function simplifies the learning problem by effectively converting a reinforcement learning problem to a supervised learning problem through the use of bootstrapping \cite{sutton1998reinforcement}, it is still a more difficult problem than conventional supervised learning. One of these additional difficulties is that the policy-dependent rewards introduces a concept drift \cite{gama2013evaluating}. This introduces the risk of unstable oscillations, but is generally solvable at the cost of slower learning if the learning rate is set sufficiently small in the beginning. This ill conditioning of the problem has been considered one of the main challenges of reinforcement learning.

\vspace{-0.4cm}

\subsection{Catastrophic forgetting}

A different problem which recently got more attention is catastrophic forgetting \cite{mccloskey1989catastrophic}. This problem is most commonly described in an online unsupervised Hebbian learning task, where the ability to retrieve previously stored patterns is lost as we update weights in the training of new patterns.This catastrophic forgetting of the original patterns takes place even when parameter space is more than sufficient to store both sets of patterns and is a consequence of the limited mixing of input objects. 

Simply alternating between the new and old patterns group with a sufficiently low learning rate would in theory solve the problem, but this has a potential impact on the convergence rate and requires explicit memorization of all training patterns. In order to minimize the effect on convergence rate, we would like to maximize the mixing of the presented inputs.

\section{Catastrophic forgetting in reinforcement learning}
The online nature of reinforcement learning means that catastrophic forgetting is a key bottleneck. There are two principal non-sharpening approaches to the catastrophic forgetting problem suggested in literature: rehearsal and pseudorehearsal. In addition, other methods based on sparse representations \cite{french1992semi} \cite{coop2013ensemble} have been used less frequently. This latter approach has a theoretical downside in its negative impact on the ability to generalize, but has shown at least mixed results and is possible to use in conjunction with the rehearsal methods.

\vspace{-0.3cm}

\subsection{Rehearsal approaches}

The first and most straight-forward principal approach for mitigating catastrophic forgetting is rehearsal \cite{ratcliff1990connectionist}, \cite{hinton1987using}. A rehearsal strategy simply stores a subset of all previous experiences in a buffer. When a new pattern is presented, this pattern is combined with several patterns from the buffer in order to form a learning batch with good mixing. There are several possible heuristics for selecting patterns for rehearsal.

The importance of catastrophic forgetting in reinforcement learning was identified early. Lin introduced the term Experience Replay \cite{lin1993reinforcement} for referring to the use of rehearsal strategies in the reinforcement learning setting. Such rehearsal has shown very promising results in robotics \cite{adam2012experience} and on more complex environments, such as Deep Q-learning for playing Atari games \cite{mnih2013playing}.

\vspace{-0.4cm}

\subsection{Pseudorehearsal approaches}

A second principal approach to solving catastrophic forgetting is pseudorehearsal \cite{robins1995catastrophic}, which does not require explicit storage of patterns. Instead, it uses a two-step process where generative models are learnt alongside with the main task. These generative models create pseudopatterns, which are combined in batches with real patterns for training the agent.

An interesting questions is whether these generated approximations of the real data are sufficiently accurate in practice to reduce forgetting. Remarkably, even extremely crude generative models have proven highly effective. In the original work in this area by \cite{robins1995catastrophic}, pure noise fed to the network was able to almost completely eliminate catastrophic interference. The argument of the authors was that, although the input is completely random, the activation distributions in deeper levels of the network will be representative of the learnt input data.

An analytical approach by Frean and Robins \cite{frean1999catastrophic} in single perceptrons suggest an alternative explanation for the surprising efficiency of random pseudopatterns. They suggest that the pseudopatterns approximate the mean of the input. Training on this mean of the input leads to decorrelation of the input patterns, which in high dimensional inputs makes the different patterns' weight updates orthogonal to each other. In addition, they demonstrated that using this mean directly was at least efficient as generating pseudopatterns. Further work in this direction was done in a thesis by Goodrich \cite{goodrich2015neuron}, where some of these results where expanded to multilayer perceptrons.

Regardless of the reason for such networks, pseudorehearsal methods have been demonstrated to significantly decrease and almost completely eliminate the catastrophic forgetting in unsupervised learning \cite{robins1995catastrophic}, supervised learning \cite{ratcliff1990connectionist} and reinforcement learning \cite{baddeley2008reinforcement}. It is interesting to note that the results of Baddeley suggest that the widely studied ill conditioning might not be the main bottleneck of reinforcement learning after all. Instead, their results indicate that the catastrophic forgetting is the main bottleneck for reinforcement learning problems.

\vspace{-0.5cm}

\subsubsection{Pseudorehearsal algorithms}
For testing pseudorehearsal approach we used two different pseudorehearsal types and the online learning with one backpropagation step as an example of learning without pseudorehearsal. One algorithm is based on correcting of the weight updates, other is a batch-backpropagation learning.

The first pseudorehearsal algorithm is the one used by Frean and Robins \cite{frean1999catastrophic} with a simplified weighting equation and changed for non-linear neural network inner assignments. The idea behind algorithm is to generate pseudoset, feed it through the network and save activations on each neuron for every pseudoexample. Then the agent is learned online, but when the real example fed to the network we use equation\\

$\Delta w_i = err_{b_i} \frac{1} {pr} \sum_{j=1}^{pr} \frac{ b_i x_{ij}\cdot x_{ij} - x_{ij} x_{ij}\cdot b_i } {b_i \cdot b_i x_{ij}\cdot x_{ij} - b_i\cdot x_{ij} b_i \cdot x_{ij}} $\\

to update $w_i$ - weights at the $i^{th}$ layer, where $err_{b_i}$ - vector backpropagation errors of the learned example at the $i^{th}$ layer, pr - size of pseudoset, $b_i$ - vector of activations of the learned example when fed forward through the network and $x_{ij}$ is vector of activations of the $j^{th}$ pseudoset on the $i^{th}$ layer.

The second one - is straight-forward using pseudosets in batch backpropagation learning - we generate set of pseudoexamples, feed it through the network, save the network outputs as the targets and then create a matrix of feature vectors where first vector is real example, others are pseudovectors, and matrix of targets where first vector is target for the real example and others are saved earlier networks outputs on pseudoset, then each time we learn agent on the whole set. 

\vspace{-0.4cm}

\subsection{Biological forgetting} 

An interesting particular case of catastrophic forgetting problem is learning in the human brain. Dual network models were initially inspired by biological learning \cite{mcclelland1995there}. As a consequence of promising experimental results of such networks, pseudorehearsal was indeed found to be the most plausible explanation for the otherwise cryptic need for dual learning systems in the brain \cite{robins1999consolidation}. 

More biologically detailed extensions of these models have recently been explored by Hattori, where they again showed excellent improvements on the ability to store information \cite{hattori2014biologically}.

The pseudorehearsal approach also contributes to the urgent need for new biological plasticity rules in large scale neurosimulation and especially for their developmental varieties (e.g. BioDynaMo \cite{BreitwieserBMJK16}). Real full scale brain simulation is approaching, but we are still lacking even a basic understanding of the role dreams play in the learning process. This despite the fact that sleep stages are of considerable length and evident in even the simplest of biological neural networks.

\section{Experimental design}

We will reevaluate the analytic results of Frean and Robins \cite{frean1999catastrophic} in a real reinforcement learning task. We evaluate and compare two algorithms for pseudorehearsal on a pole balancing task using Q-learning with function approximation. Further, we will study the effect of sparsity in fulfilling the requirement for a high dimensional input space these algorithms relied on. Our agent is a classic Q-learning agent with $\epsilon$-greedy policy using a feed-forward-backpropagation neural network as function approximator, discounted factor of agent is 0.9. The environments used for training is the single-pole balancing cart. 

\subsubsection{Observation}
Two different observations are used for experimental comparison. The first observation type given to the agents constitute a fully observable MDP and includes current position, velocity, acceleration of the cart, as well as the current angle, angular velocity and angular acceleration of the pole. The second observation type make the problem partially observable - here the agent knows only cart's position and pole's angle. If the cart reaches the end of track or the pole falls for angle more than predefined pole failing angle - the game is lost and the agent is gained negative reward for this task. 

We represented the observations as a feature vector by two different methods. The first method was a representation of input values where the $i$-th observation value sets the $2*i$-th feature if it is positive or on $2*i+1$-th if it is negative. The second method was to use sparse unary vectors where feature vector is concatenation of parameter vectors, similar to a table. Each of parameter vectors consisted of elements associated with discrete values inside the range possible for each parameter - $[-20; 20]$ for linear parameters, $[-60; 60]$ for angular. All the elements of the vector were set to zeros except two - the element associated with the rounded value of the parameter was set to one, and the next element is set to the fractional part of this parameter.\\

\vspace{-1cm}

\subsubsection{Performance metric}
Agent tries to balance pole or poles as long as it can for 5000 tries, for two sets of parameters we also made an averaged variants where 10 iterations of this 5000 tries are averaged to make sure that convergence tendency is reproducible and not a set of random successful moves. We also have results for fully random policy to compare with.\\

\vspace{-1cm}

\subsubsection{Parameter settings}
The task was repeated with different sizes of pseudoitem batches, with different numbers of iterations between reinitialization of the pseudosets and with different learning rates. The learning rates used were 0.1, 0.01 and 0.001. The discount factor was set to 0.9. The sizes of pseudosets were 10, 30, 50 and 100 pseudoitems, respectively. We resample a new set of pseudoitems after every 1, 10 or 100 runs. Parameters are chosen to define influence of size of the pseudoset and frequency of it's reinitialisation on learning. We try to cover a wide range without trying all the possible values. And we decrease parameters in close to geometrical progression to see if the influence is logarithmic. For 30 and 50 item pseudorehearsal batches we also tried 30 and 50 reinitialisation gaps to increase coverage.
\\
\vspace{-1cm}

\subsubsection{Performance metrics}
We did not stop learning after an agent reaches satisfactory result, because the continued learning contains cases of catastrophic forgetting which we would like to explore. As we had a very short learning time during which the performance increases followed by long row of tries with unstable behavior, we evaluate the efficiency of different approaches by measuring mean and median number of steps per try for each approach and compare these two numbers. 

$Mean > median$ indicates that some agent's tries were highly effective and agent balanced pole for a long time, while the most of runs were weak, so no convergence occurred or the influence of catastrophic interference is too high to handle needed weights for a long time. $Mean<median$ shows that agent successfully converge to some optimal policy, and it's policy is stable but some tries are failed so bad that it affected the whole picture, so in this case we can see successful learning, strong influence of catastrophic forgetting and successful avoidance of this influence. $Mean \approx meadian$ means that catastrophic forgetting and its avoidance has nearly the same influence. 
\begin{figure}[!htb]
\includegraphics[width=1\textwidth]{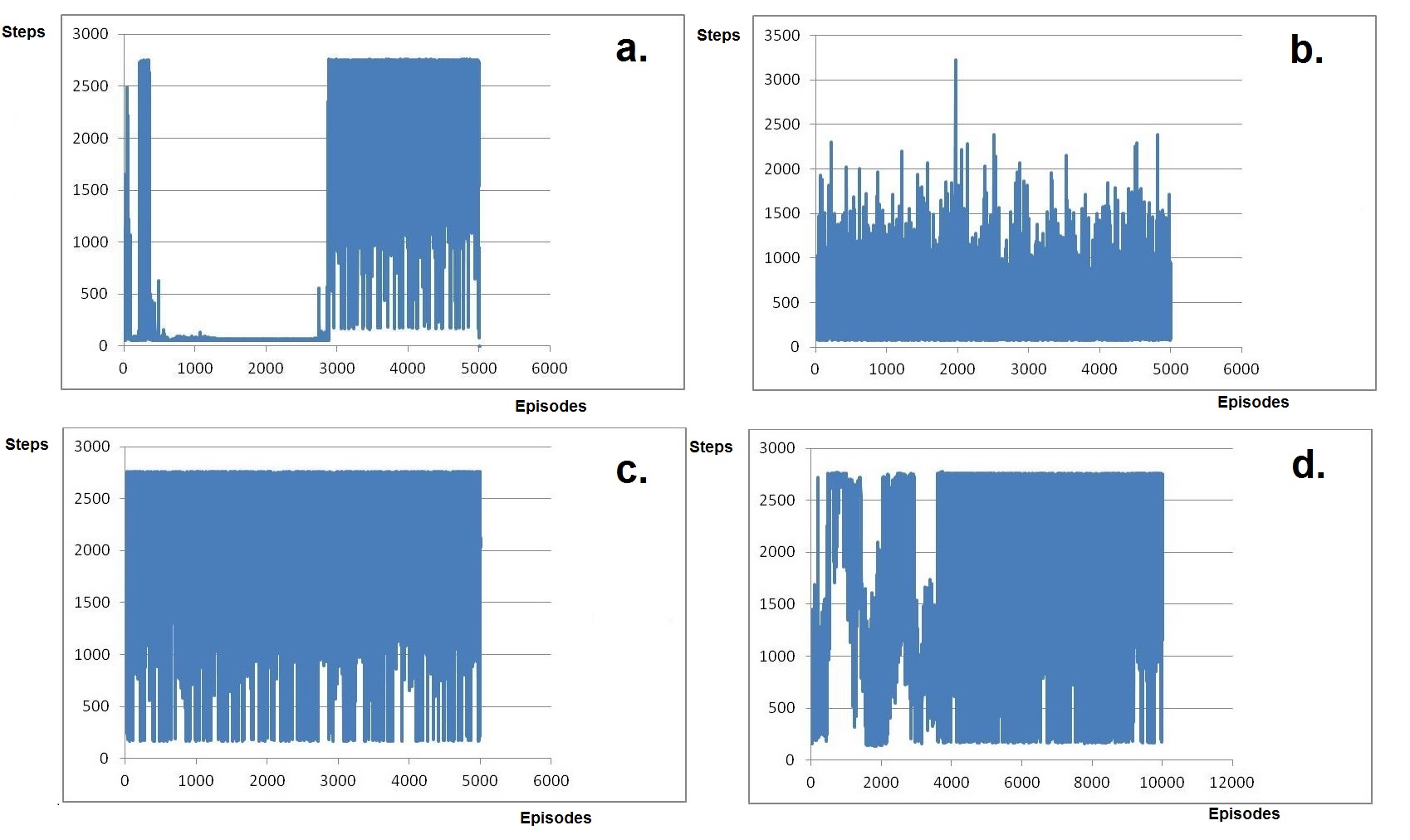}
\caption{Examples: a. median $\ll$ mean; b. median $<$ mean; c. median $>$mean; d. median $\approx$ mean}
\label{fig:figure1}
\end{figure}
Results were averaged over ten runs when training times allowed and other cases we presents results over single runs.

\subsection{Results}
Results for all observations show different learning for different cases, while averaged approach and comparison with the random agent assure us that the learning has place and in case without pseudorehearsal it depends on learning rate mostly. For the agents using pseudorehearsal learning depends on sets of parameters - learning rate, pseudoset size and relearning gap. Some of them can make agent to balance pole for significantly larger time than a random run, while the others perform the same, or worse, or even worse than a random agent. 

For both used metrics we discovered a roughly bell-shaped graph of dependencies for each of used learning rates and for each of used techniques. All pseudorehearsal approaches have different sets of parameters providing optimal learning, a suboptimal and worse - in case of parameters a little different from the optimal ones. Further away from optimal parameters agents started to diverge, e.g. tried to drop the pole about four times faster than if it would use random policy, etc. 

Both Frean-Robins and batch approaches perform similarly in for this observation, but their optimal parameters differ for the same learning rates. All this results are summarized in table ~\ref{table:mytable1} for MDPs and table ~\ref{table:mytable2} for partially observable MDPs - POMDPs.\\

\begin{table}[!htb]
\includegraphics[width=0.95\textwidth, inner]{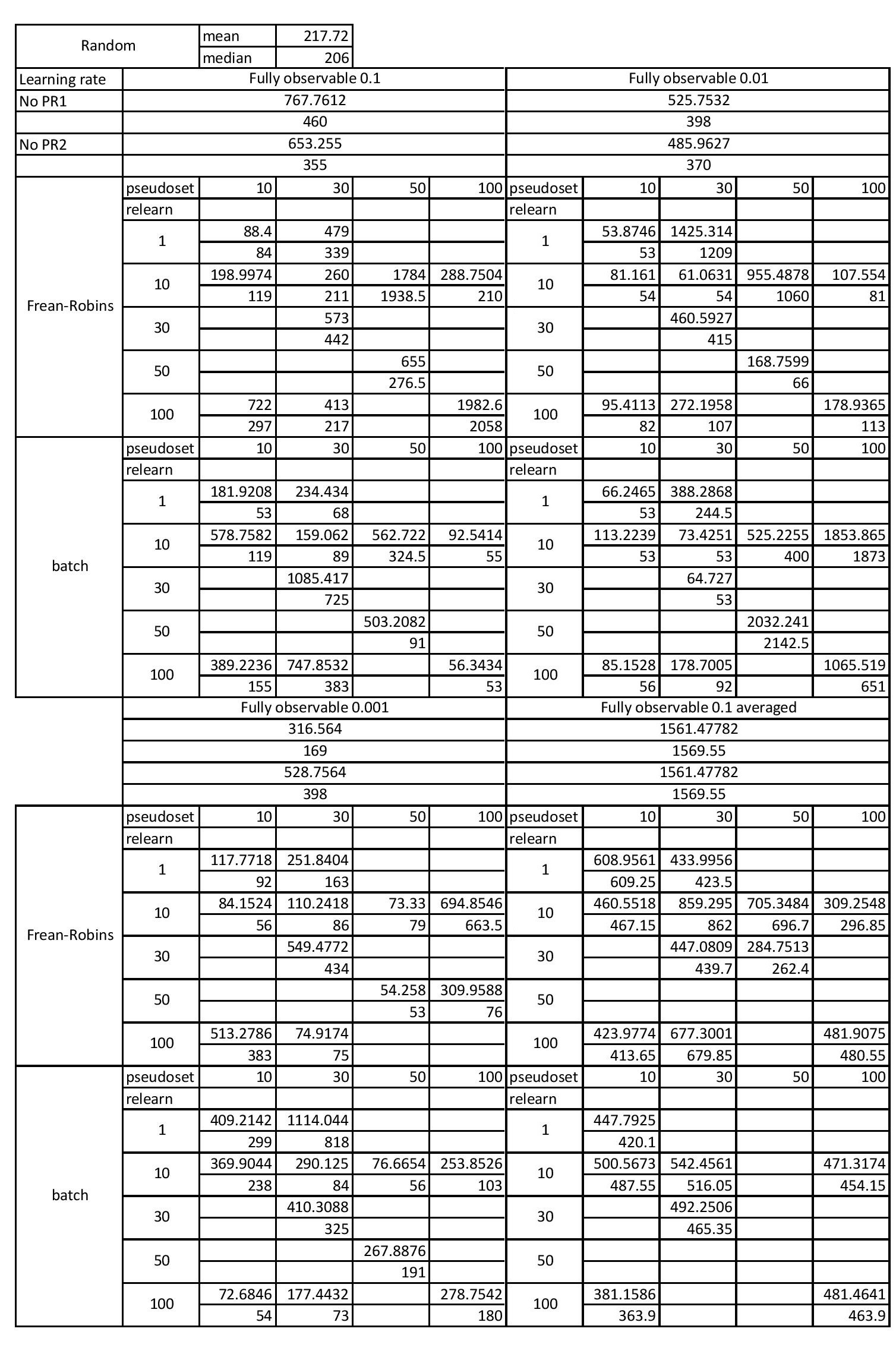}
\caption{MDP results overall}
\label{table:mytable1}
\end{table}

\subsection{MDP}
For the cases of fully observable MDP, where agent knows anything about the current state of the pole cart and the pole the learning time when the agent's performance goes from some initial random to the final one is very short, agent quickly converges at some number of steps it can balance and most of its next moves holds around this result with some deviations, sometimes very large, caused by catastrophic forgetting.
\begin{figure}[!htb]
\includegraphics[width=1\textwidth, inner]{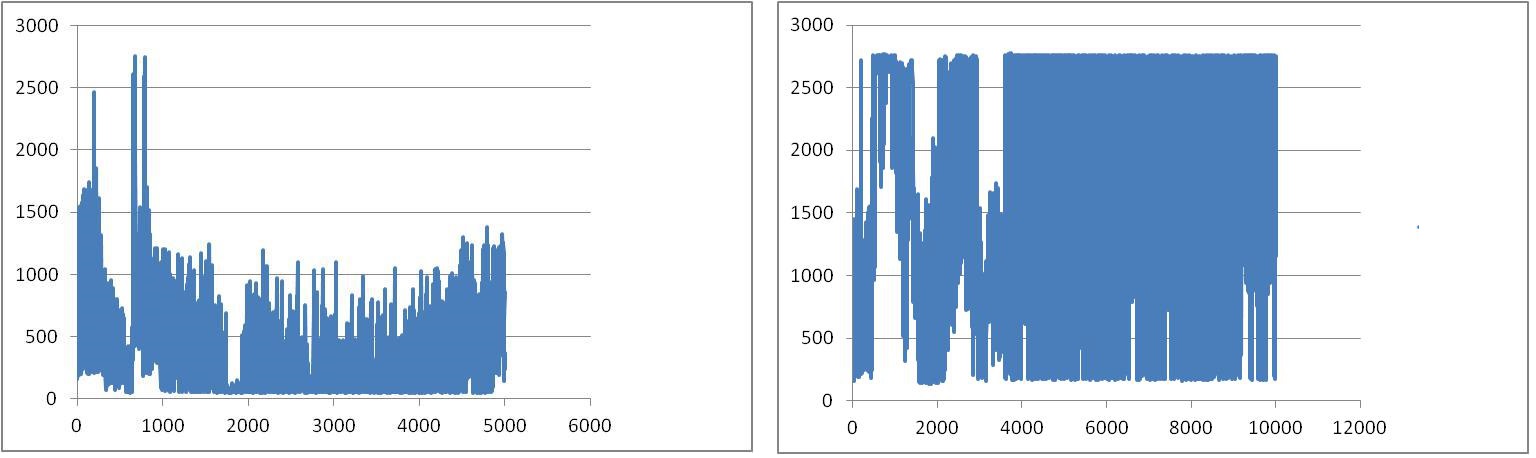}
\caption{left plot - learning rate = 0.001, batch learning, pseudoset size =10, relearning gap = 10; right - learning rate = 0.01, pseudoset size = 100, relearning gap = 100}
\label{fig:figure2}
\end{figure}
If some learning case makes agent significantly change its behaviour - change is as quick as initial learning and on graph looks like immediate change of performance.

\subsection{POMDP}
Partial observability tends to suffer less from forgetting, possibly because each part of the smaller space are more frequently visited. Pseudorehearsal has a more significant impact here: although it shares the same optimal-suboptimal-worst sets of parameters, optimal ones further decrease the number of runs needed to learn and decrease influence of the catastrophic forgetting: if the agent in current set of parameters doesn't diverges to the worst possible case - it's median for all runs is always higher then the mean, indicating a relatively stable behaviour after training. 
\begin{table}[!htb]
\includegraphics[width=1.0\textwidth, left]{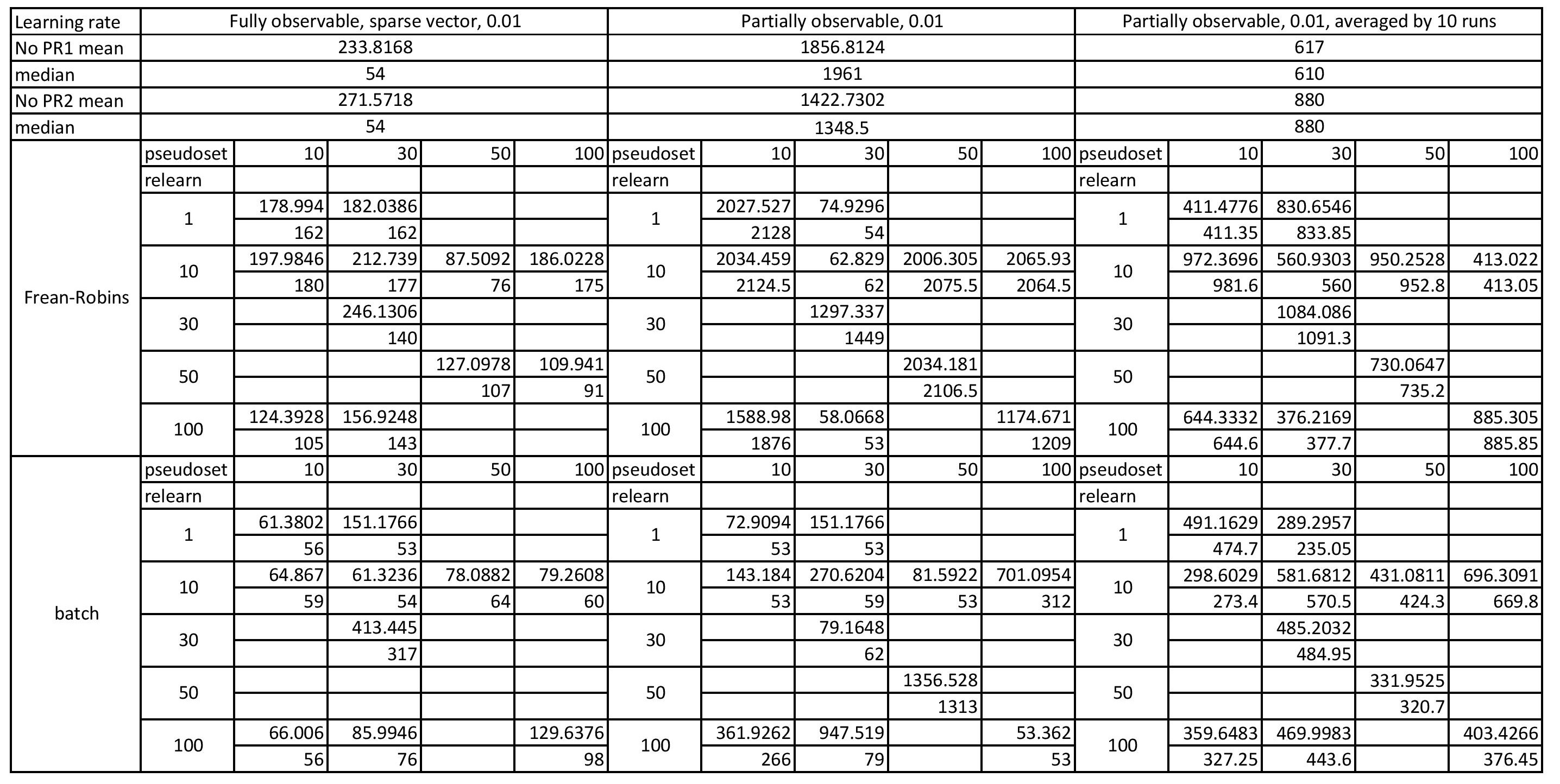}
\caption{POMDP results overall}
\label{table:mytable2}
\end{table}
On the other hand, agents in POMDPs agents that can't converge change their policy more frequently than agents in MDPs, and while the agent in fully observable MDP has a minor chance to reach good performance after it reached a suboptimal solution, agent in POMDP easily changes its policy both ways. 
\begin{figure}[!htb]
\includegraphics[width=1\textwidth, inner]{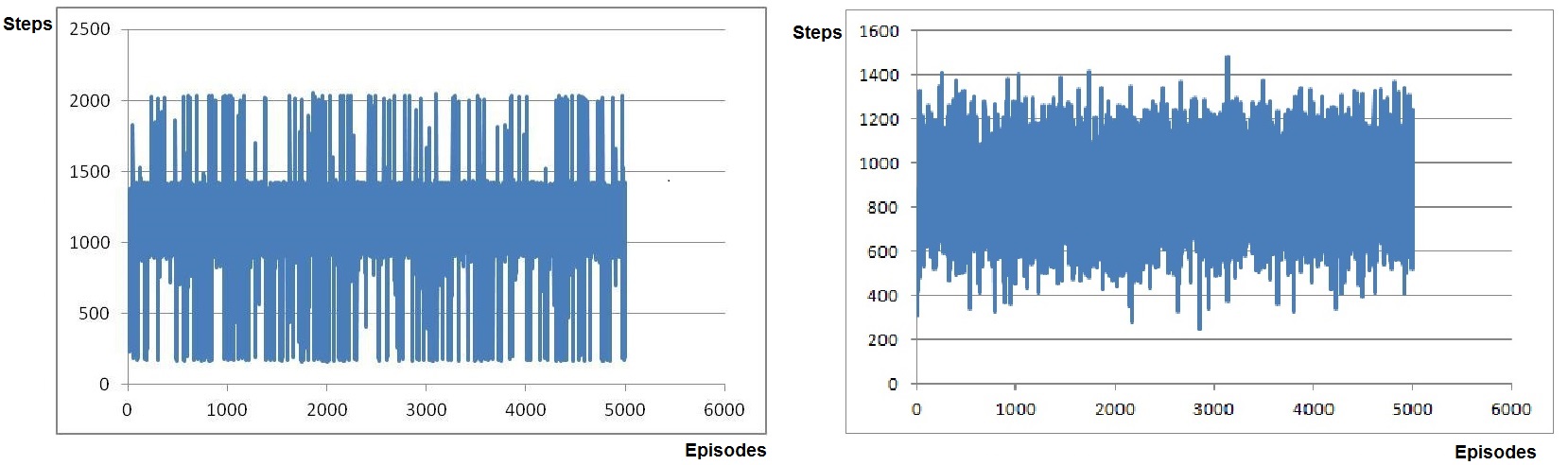}
\caption{results of running in POMDP with pseudoset size 100 and relearning gap 100, single run on the left plot and averaged by 10 runs on the right one}
\label{fig:figure3}
\end{figure}

The pseudorehearsal approach taken from Frean and Robins decreases this serious context switches and all the agents using this type of pseudorehearsal show nearly the same performance during the all runs, holding around some value with occasional deviations, results much better or worse can be met, but they are rare compared to the results close to this mean.\\
\begin{figure}[!htb]
\includegraphics[width=0.5\textwidth, center, inner]{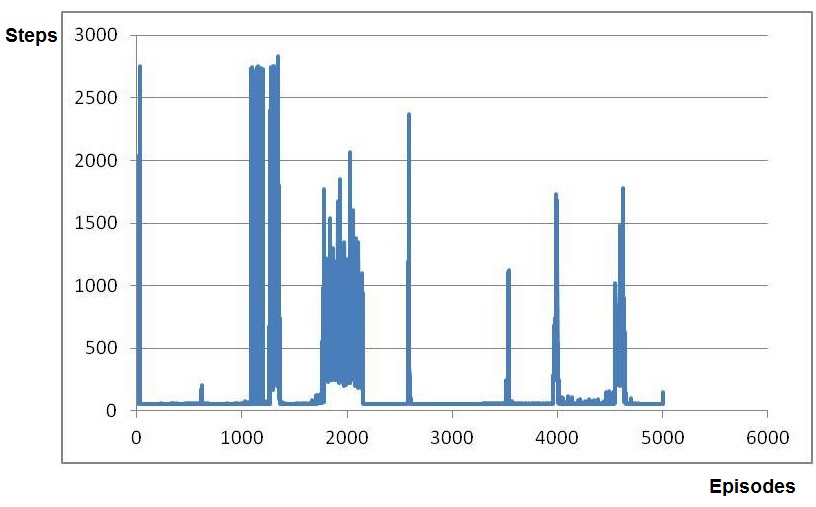}
\caption{Example of diverged agent changing policy: batch pseudorehearsal, pseudoset size = 30, no relearning gap - new pseudoset generated after each run}
\label{fig:figure4}
\end{figure}
Stability is maintained for all sets of parameters, while the mean value can differ. Different effect is caused by batch-backpropacation learning using pseudosets: picture of agent's performance is the same as in learning without pseudorehearsal, but efficiency switching occurs only after reinitialization of the pseudoitem vector. While same pseudorehearsal parameters may lead to different agent's behaviour, the efficiency of each set of parameters evaluated by averaging for ten iterations shows that some sets make agent to increase it's performance during the time, while the others do not. \\

\vspace{-0.8cm}

\section{Discussion and conclusions}
The experiment has shown us that pseudorehearsal can deal with catastrophic interference, but it has its own effects which in some cases cause divergence that worsen performance, so this tool should be used carefully and the parameters - learning rate, pseudoset size and relearning frequency have to be chosen properly to guarantee high performance on the current task.

For the fully observable MDPs pseudorehearsal decreases influence of the catastrophic forgetting if the optimal parameters for the task are known. In the best cases, optimal performance was reached quickly with pseudorehearsal, but the further parameters from the optimal, the worse performance was. In the case if modeling this environment might be too complex - optimal parameters can be defined only empirically before starting learning, which may be unacceptable if the cost of mistake is high. 

For partially observable environments all the problems met by fully observable ones remain the same and some additional effects were noted: agent's policy doesn't only converge, diverge or stay random, but also converge to some value with the majority of tries having results in a some range around this value, and strongly deviating runs are more rare and separated by wide gaps of convergence. Another notable effect of the pseudorehearsal in POMDP agents with both pseudorehearsal cases is a significant decrease of the number of steps to converge to the number of steps needed by fully observable agent. If an agent in fully observable environment can converge it converges at about 20-30 runs as with pseudorehearsal, so without, if agent in partially observable environment can converge it converges at 100-250 runs without pseudorehearsal and at 10-30 runs with pseudorehearsal.
\begin{figure}[!htb]
\includegraphics[width=1.0\textwidth, inner]{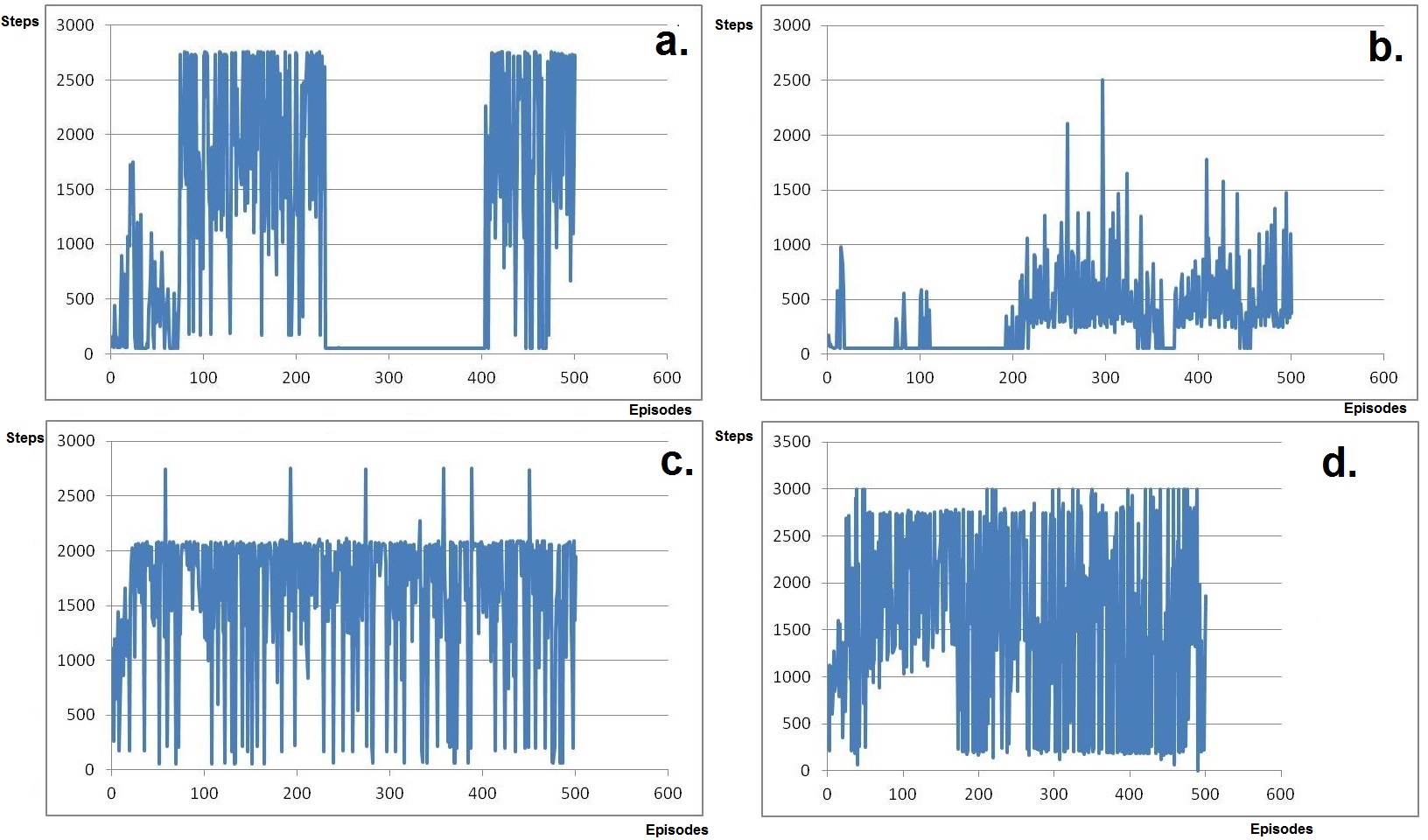}
\caption{a., b.500 first stps for two agents without pseudorehearsal; c. one agent with Frean-Robins' pseudorehearsal with pseudoset size = 10 and relearning gap = 100; d. one agent with batch pseudorehearsal with pseudoset size = 100 and relearning gap = 10}
\label{fig:figure5}
\end{figure}

Pseudorehearsal is known to be a powerful tool for improving performance of supervised learning agents. We have shown that it can be useful to assist learning even in relatively quickly mixing continuous reinforcement learning tasks, if parameters are chosen correctly. Pseudorehearsal reduces this forgetting effect and maintains stable solutions for longer. While pseudorehearsal may strongly improve agent's performance and accelerate learning, empirical defining of the optimal pseudoset size and relearning gap is required. One of possible extension of this research would be exploration of mathematical way to figure out this parameters. We will also explore new, more complex reinforcement learning challenges and try more advanced dual network generation of pseudoexamples.

\bibliographystyle{IEEEtran}
\bibliography{IEEEabrv,mainbib}

\end{document}